\documentclass[letterpaper, 10 pt, conference]{ieeeconf}  

\let\labelindent\relax
\usepackage{enumitem}  

\IEEEoverridecommandlockouts                              
\usepackage{xcolor}
\usepackage{graphicx}
\usepackage{enumitem}
\usepackage{placeins}
\usepackage{float}

\usepackage{algorithm, algorithmicx, algpseudocode}
\usepackage{siunitx}
\usepackage{booktabs}

\definecolor{nnmodulecolor}{RGB}{255, 235, 205}

\title{\LARGE \bf
SPAQ-DL-SLAM: Towards Optimizing Deep Learning-based SLAM for Resource-Constrained Embedded Platforms\vspace{-5pt}}

\author{Niraj Pudasaini$^{1}$, Muhammad Abdullah Hanif$^{1}$, Muhammad Shafique$^{1}$
\thanks{$^{1}$ All authors are affiliated with New York University Abu Dhabi (NYUAD), UAE 
        {\tt\small \{np2289, mh6117, muhammad.shafique\}@nyu.edu}}\vspace{-20pt}
}

\begin{document}

\maketitle

\thispagestyle{empty}
\pagestyle{empty}

\begin{abstract}

Optimizing Deep Learning-based Simultaneous Localization and Mapping (DL-SLAM) algorithms is essential for efficient implementation on resource-constrained embedded platforms, enabling real-time on-board computation in autonomous mobile robots. This paper presents SPAQ-DL-SLAM, a framework that strategically applies Structured Pruning and Quantization (SPAQ) to the architecture of one of the state-of-the-art DL-SLAM algorithms, DROID-SLAM, for resource and energy-efficiency. Specifically, we perform structured pruning with fine-tuning based on layer-wise sensitivity analysis followed by 8-bit post-training static quantization (PTQ) on the deep learning modules within DROID-SLAM. Our SPAQ-DROID-SLAM model, optimized version of DROID-SLAM model using our SPAQ-DL-SLAM framework with 20\% structured pruning and 8-bit PTQ, achieves an 18.9\% reduction in FLOPs and a 79.8\% reduction in overall model size compared to the DROID-SLAM model. Our evaluations on the TUM-RGBD benchmark shows that SPAQ-DROID-SLAM model surpasses the DROID-SLAM model by an average of 10.5\% on absolute trajectory error (ATE) metric. Additionally, our results on the ETH3D SLAM training benchmark demonstrate enhanced generalization capabilities of the SPAQ-DROID-SLAM model, seen by a higher Area Under the Curve (AUC) score and success in 2 additional data sequences compared to the DROID-SLAM model. Despite these improvements, the model exhibits performance variance on the distinct Vicon Room sequences from the EuRoC dataset, which are captured at high angular velocities. This varying performance at some distinct scenarios suggests that designing DL-SLAM algorithms taking operating environments and tasks in consideration can achieve optimal performance and resource efficiency for deployment in resource-constrained embedded platforms.
\end{abstract}

\section{INTRODUCTION}
Simultaneous Localization and Mapping (SLAM) is a technique that enables an agent to map an unknown environment while simultaneously tracking its location. SLAM is used for robotic autonomy and is an intrinsic component of embodied AI systems for navigation \cite{zhu2021deep} \cite{ZHANG2022105036}. As embodied agents are increasingly deployed in complex, dynamic real-world scenarios, the demand for accurate, robust, yet computationally efficient onboard SLAM systems has grown significantly \cite{Cadena_2016}.

Current SLAM approaches can be categorized into classical and deep learning-based methods. Classical SLAM, exemplified by ORB-SLAM3 \cite{ORBSLAM3_TRO}, relies on handcrafted feature detection and matching algorithms. While computationally efficient, these methods often struggle in complex, dynamic environments \cite{bujanca2021robust}. In contrast, DL-SLAM systems, such as DROID-SLAM \cite{teed2021droid}, Rover-SLAM \cite{xiao2024realtimerobustversatilevisualslam}, GO-SLAM \cite{zhang2023goslamglobaloptimizationconsistent}, and DVI-SLAM \cite{peng2024dvislamdualvisualinertial}, integrate deep neural networks (DNNs) to enhance the perception and interpretation of the environment, leading to significant improvements in accuracy, robustness, and generalization. Specifically, DROID-SLAM shows a 43\% reduction in error over ORB-SLAM3 on the EuRoC dataset and an 83\% reduction on the TUM-RGBD dataset for the sequences where ORB-SLAM3 is successful \cite{teed2021droid}.

However, the computational demands of DL-SLAM systems (see Table \ref{tab:hardware_requirements}) pose a significant challenge for on-board computation in mobile robots. These requirements often exceed the processing capabilities of embedded platforms commonly used in mobile robots (see Table \ref{tab:mobile_robots}).

\begin{table}[h!]
\centering
\caption{Hardware Configurations Used in Testing DL-SLAM Systems}
\label{tab:hardware_requirements}
\resizebox{\columnwidth}{!}{%
\begin{tabular}{@{}ccc@{}}
\toprule
\textbf{DL-SLAM System} & \textbf{GPU Model (CUDA Cores)} & \textbf{Memory Size (GB)} \\
\midrule
DROID-SLAM \cite{teed2021droid} & $2 \times \text{NVIDIA RTX 3090}$ ($2 \times \text{10,496}$) & $2 \times \text{24}$ (GDDR6X) \\
GO-SLAM \cite{zhang2023goslamglobaloptimizationconsistent} & NVIDIA RTX 3090 (10,496) & 24 (GDDR6X) \\
Rover-SLAM \cite{xiao2024realtimerobustversatilevisualslam} & NVIDIA RTX 3080 (8,704) & 10 or 12 (GDDR6X) \\
DVI-SLAM \cite{peng2024dvislamdualvisualinertial} & NVIDIA RTX 3090 (10,496) & 24 (GDDR6X) \\
\bottomrule
\end{tabular}%
}
\end{table}

For example, the real-time implementation of DROID-SLAM requires two NVIDIA RTX 3090 GPUs, each equipped with 10,496 CUDA cores (see Table \ref{tab:hardware_requirements}). The system divides its workload, running front-end tracking and local bundle adjustment on the first GPU, while the second GPU handles back-end global bundle adjustment and loop closure, requiring a total memory of around 24 GB \cite{teed2021droid}. In contrast, one of the most powerful embedded platforms in mobile robotics, the NVIDIA Jetson AGX Orin, has only 2,048 CUDA cores—just 9.76\% of what DROID-SLAM necessitates—and significantly less memory capacity (see Table \ref{tab:mobile_robots}). This disparity significantly challenges the deployment of state-of-the-art DL-SLAM algorithms on embedded platforms for mobile robots.

\begin{table}[h!]
\centering
\caption{Computational and Battery Capacity of Common Mobile Robots in Robotics R\&D}
\label{tab:mobile_robots}
\resizebox{\columnwidth}{!}{%
\begin{tabular}{@{}ccc@{}}
\toprule
\textbf{Robot (Category)} & \textbf{GPU Model (CUDA Cores)} & \textbf{Memory / Battery Capacity} \\
\midrule
NVIDIA JetBot (UGV) & NVIDIA Jetson Nano (128) & 4 GB / 10,000 mAh Li-Po \\
Skydio 2 (Drone) & NVIDIA Jetson TX2 (256) & 8 GB / 4,280 mAh \\
Boston Dynamics Spot (Legged) & NVIDIA Jetson Xavier (512) & 16 GB / 605 Wh \\
Clearpath Robotics Husky (UGV) & NVIDIA Jetson AGX Orin (2048) & 32 GB / 2x 10Ah, 24V Li-Ion \\
\bottomrule
\end{tabular}%
}
\end{table}

\textbf{Our Novel Contributions:} This paper presents a resource and energy-efficient optimization framework SPAQ-DL-SLAM (see Fig \ref{fig:spaqs_optimization_pipeline}), applied to the architecture of DROID-SLAM. To the best of our knowledge, this is the first implementation of neural network optimization techniques in a DL-SLAM algorithms. Our main contributions are:

\begin{figure}[H]
    \centering
    \includegraphics[width=\linewidth]{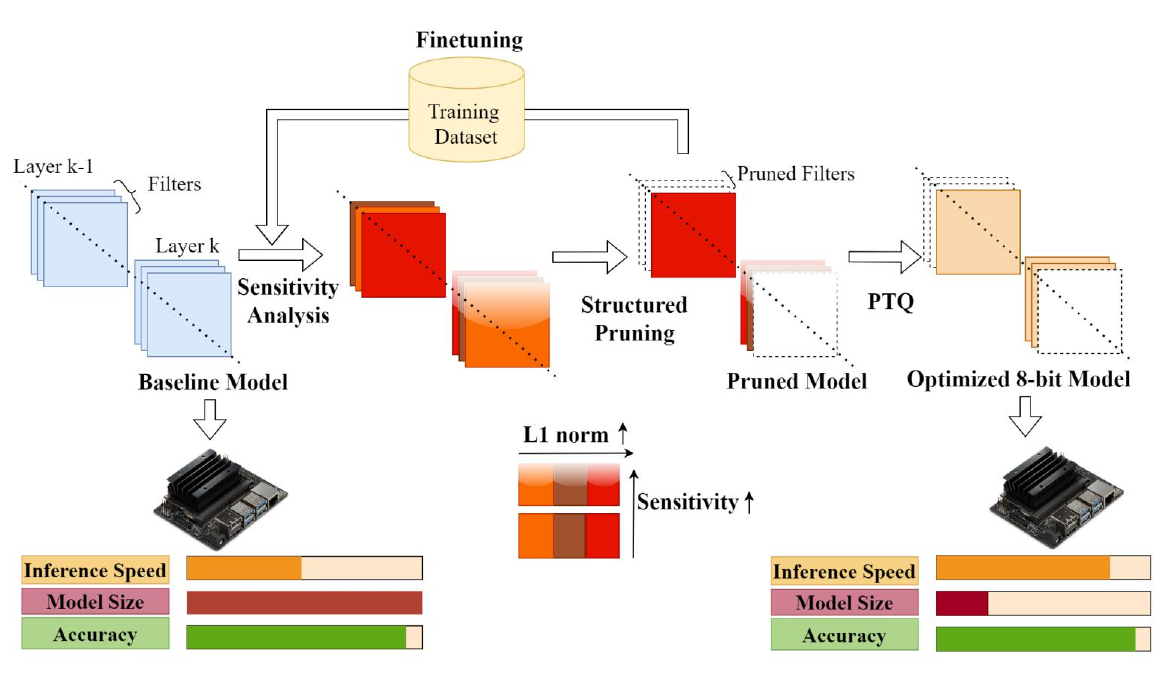}
    \caption{Pipeline for SPAQ-DL-SLAM. The original model layers (e.g., Layer k-1 and Layer k) are first analyzed for sensitivity to pruning impacts. This is followed by iterative fine-tuning and structured pruning across all sensitive layers. The model is then quantized to 8-bit using PTQ, enhancing computational efficiency while maintaining or improving performance metrics such as inference speed, model size, and accuracy.}
    \label{fig:spaqs_optimization_pipeline}
\end{figure}

\begin{enumerate}[leftmargin=*]
    \item We present SPAQ-DL-SLAM, an optimization framework for DL-SLAM systems, implemented on DROID-SLAM. This framework incorporates structured pruning and fine-tuning based on layer-wise sensitivity analysis, followed by 8-bit PTQ in the DNN modules.

    \item Comprehensive evaluations show that SPAQ-DROID-SLAM, using 20\% structured pruning and 8-bit PTQ, reduces FLOPs by 18.90\% and model size by 79.8\%. It outperforms the original DROID-SLAM baseline on the TUM-RGBD benchmark by 10.5\% on ATE metric, and demonstrates better generalization on the ETH3D SLAM benchmark, shown by a higher AUC score and success in 2 additional sequences. 

    \item Our evaluation shows that high angular velocity data, especially from EuRoC's Vicon Room sequences using MAVs, degrades SPAQ-DROID-SLAM performance, highlighting the need to tailor DL-SLAM algorithms for specific environments and tasks for optimal efficiency.
\end{enumerate}

\section{BACKGROUND AND RELATED WORK}
\subsection{DROID-SLAM}
The integration of DNNs to solve SLAM problem a shift motivated by the superiority of DNNs in various computer vision tasks, including image classification \cite{krizhevsky2012imagenet}, object detection \cite{ren2015faster}, and segmentation \cite{long2015fully}, which suggest potential for similar advancements in SLAM systems. 

DROID-SLAM exemplifies this shift in the visual SLAM domain, utilizing DNN modules in the architecture, to predict optical flow and update camera poses and pixel-wise depths through a Dense Bundle Adjustment (DBA) process. Its development shows significant achievements, surpassing previous work in accuracy and robustness, with fewer catastrophic failures when compared to the classical state-of-the-art SLAM systems. Moreover, it outperforms other Deepv2d \cite{DBLP:journals/corr/abs-1812-04605}, Deepfactors \cite{DBLP:journals/corr/abs-2001-05049}, and Tartanvo \cite{DBLP:journals/corr/abs-2011-00359} DL-SLAM systems by high-margins in TartanAir, ETH3D-SLAM, EuRoC, and TUM-RGBD visual SLAM datasets. \cite{teed2021droid}. 

\subsubsection{Main features of DROID-SLAM}
This section highlights DROID-SLAM’s core techniques.

\textbf{Full Bundle Adjustment:} 
An integral part of its system for increase of its accuracy is the full Bundle Adjustment that optimizes camera poses and 3D map points within a single framework, allowing for sensor versatility and improved geometric consistency.

\textbf{Optimization of Geometric Error:} 
DROID-SLAM optimizes geometric error directly, which contributes to its state-of-the-art flow estimates.

\textbf{Blend of Indirect and Direct Approaches:}
By leveraging the full image feature instead of sparse points, it combines the strengths of both indirect and direct methods, enhancing robustness and accuracy.

\begin{figure}[h!]
\centering
\includegraphics[width= 0.5\textwidth]{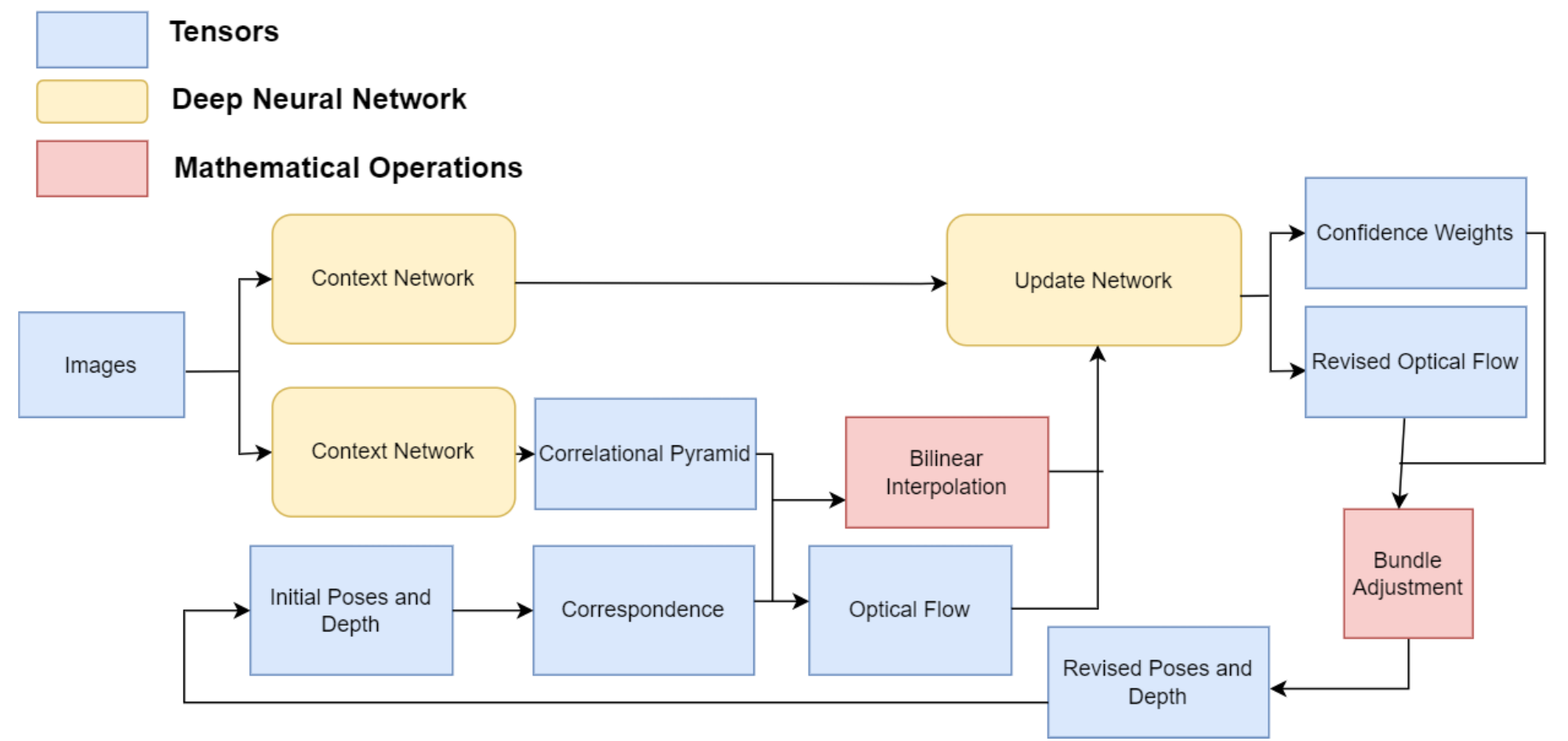}
\caption{ DROID-SLAM operational flow, highlighting key components.}
\label{fig:droidslamarch}
\end{figure}


\subsubsection{DROID-SLAM}
DNNs of DROID-SLAM are composed of three distinct neural network modules: Feature, Context, and the Update Network (see Fig \ref{fig:droidslamarch}).

\begin{figure}[h!]
    \centering
    \includegraphics[width=0.35\linewidth]{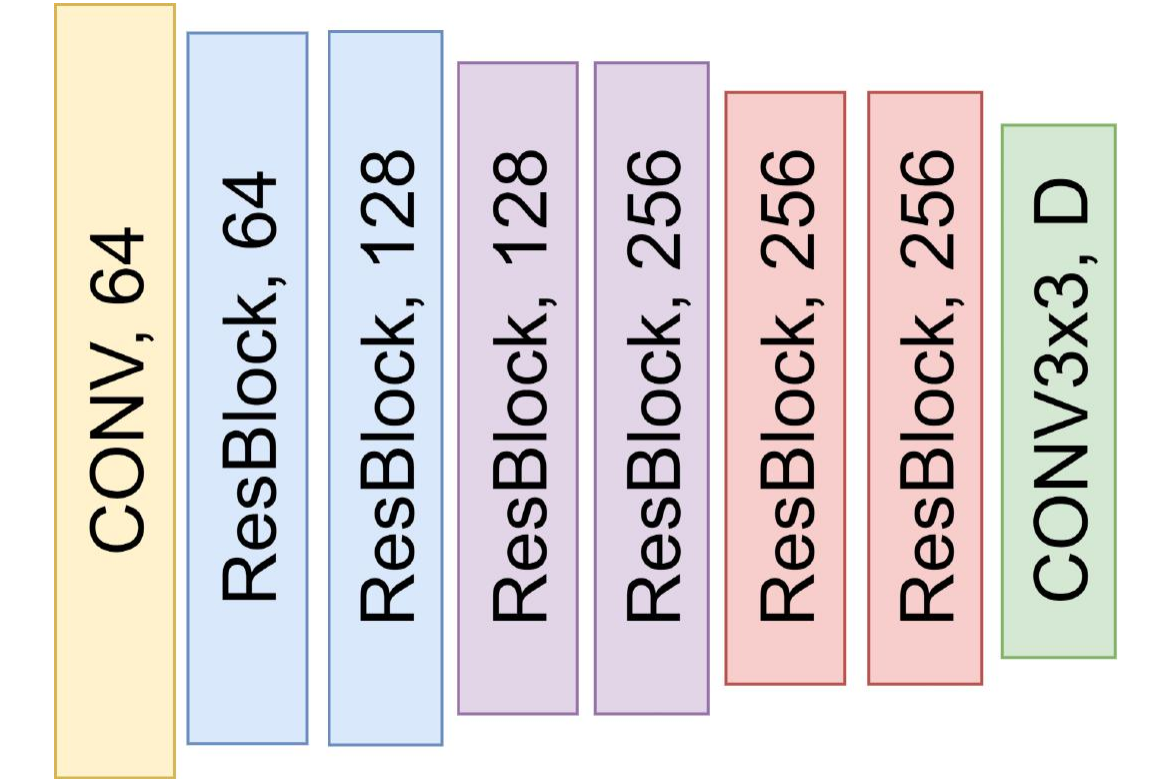}
    \caption{Feature and Context Network Architecture: Utilizes six residual blocks for 1/8 resolution feature extraction. The feature encoder uses instance normalization, producing \(D=128\), whereas the context encoder, which omits normalization uses \(D=256\). Figure is adapted from \cite{teed2021droid}.}
    \label{fig:feature_context_encoders}
\end{figure}

\textbf{Feature Network and Context Network:} Both feature and context networks share a structure composed of 18 convolutional layers, with feature network having instance normalization across its layers. It's design includes 6 residual and 3 downsampling blocks, with feature maps of dimension 128 (see Fig \ref{fig:feature_context_encoders}). Context network follows the same layout but omits instance normalization, resulting in feature maps with a dimension 256. This design choice helps the networks to capture and correlate essential features across consecutive frames for robust optical flow and depth estimation. 

\begin{figure}[h!]
    \centering
    \includegraphics[width=0.5\linewidth]{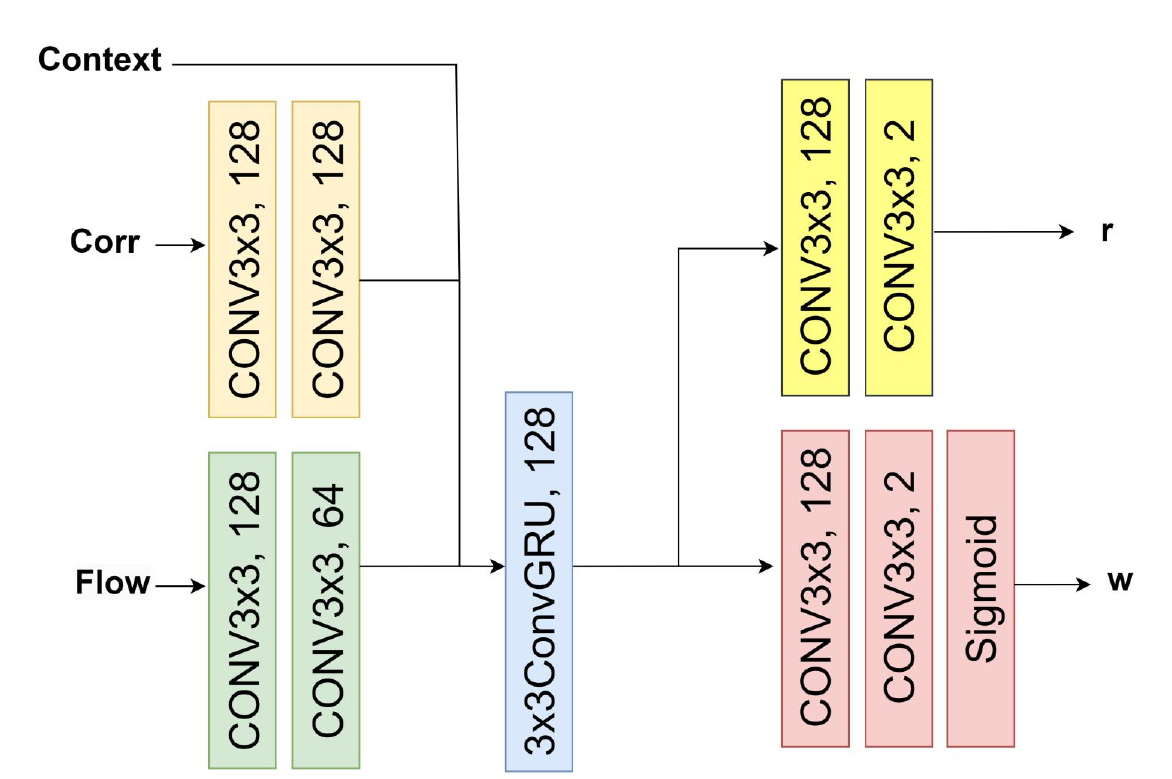}
\caption{Update Network: Iteratively integrates context, correlation, and flow features into a GRU, enabling prediction of revisions (\(r\)) and confidence weights (\(w\)) from the evolving hidden state \cite{teed2021droid}. Figure is adapted from \cite{teed2021droid}.}
    \label{fig:update_operator}
    \end{figure}

\textbf{Update Network:} The refinement of optical flow and depth estimation is performed by the update network, which incorporates a ConvGRU module alongside multiple convolutional layers (see Fig \ref{fig:update_operator}). This network contains 19 convolutional layers across its submodules. 

DNNs in DROID-SLAM make significant use of convolutional neural networks (CNNs) (see Table \ref{tab:computational_profile}).
\begin{table}[H]
\centering
\caption{Summary of DROID-SLAM DNN Computational Profile}
\resizebox{0.3\textwidth}{!}{
\begin{tabular}{l r}
\hline
\textbf{Metric} & \textbf{Value} \\
\hline
Total Model Parameters & 4.00 M \\
Total CNN Parameters & 3.94 M \\
CNN Parameters in \% & 98.5\% \\
FLOPs (CNN Parameters) & 4.64 B \\
\hline
\end{tabular}
}
\label{tab:computational_profile}
\end{table}


\subsection{Pruning}
The principle behind network pruning is to identify and eliminate the weights or filters in a network that are less important, thereby streamlining the model for faster inference and lower memory requirements. The origins of network pruning can be traced back to early works \cite{lecun1989optimal} \cite{hanson1989comparing} \cite{hassibi1993second} that explored the idea of minimizing network complexity to improve generalization and reduce overfitting.

The paper \cite{han2015deep} introduced a novel approach that applied pruning to state-of-the-art CNN models, demonstrating that it was possible to significantly reduce the number of parameters without losing accuracy. Their method involved a three-step process: initial training to determine the importance of connections, pruning of those deemed least important, and subsequent retraining to fine-tune the sparse network. This approach allowed for reductions in model parameters by 9× for AlexNet and 13× for VGG-16, showing the potential of pruning in modern deep learning applications.

\begin{figure}[h!]
\centering
\includegraphics[width=0.45\textwidth]{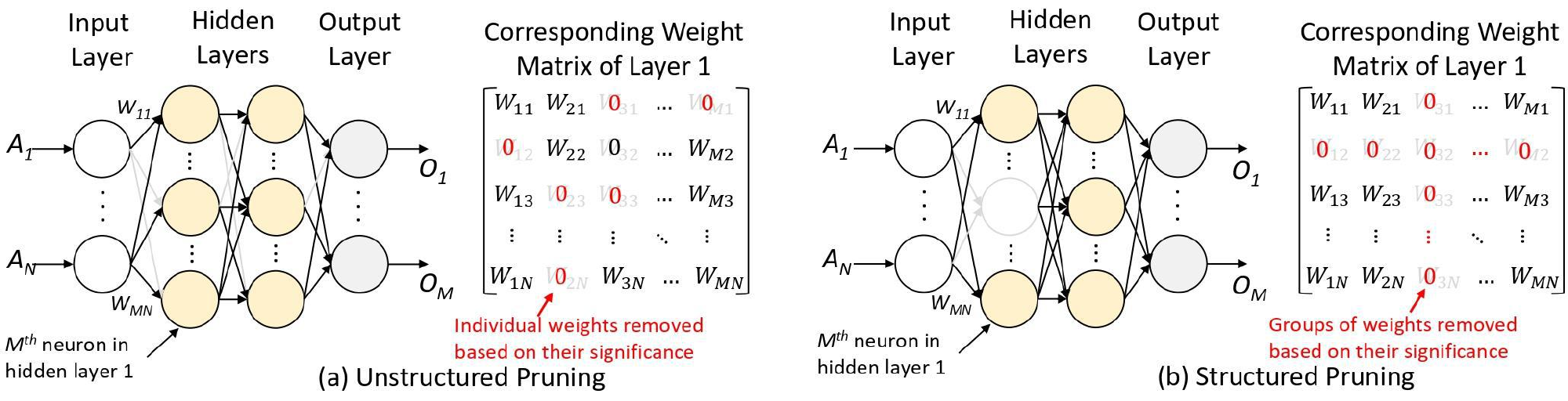}
\caption{(a) Unstructured pruning involves pruning individual weights in a neural network, leading to a sparse architecture. (b) Structured pruning targets entire filters or channels for removal, which can maintain the dense structure of the network while reducing the number of filters or channels}
\label{fig:pruning_types}
\end{figure}
In the context of network pruning, there are two main approaches: unstructured (a.k.a. fine-grained pruning) and structured pruning (a.k.a. coarse-grained pruning) as illustrated in the  Figure \ref{fig:pruning_types}. Unstructured pruning focuses on removing individual weights across the network, leading to a sparsely connected architecture. In contrast, structured pruning targets entire filters or channels for removal, preserving the dense structure of the network but with fewer filters or channels. Structured pruning is particularly relevant for CNNs, as it aligns with the architecture's inherent structure, facilitating easier implementation on hardware without specialized support for sparse computations \cite{hanif2022crosslayer}. Numerous studies have shown that structured pruning can effectively reduce the complexity of CNNs while maintaining, or even improving, model accuracy and inference speed \cite{li2016pruning} \cite{luo2017thinet}.

Given the non end-to-end DNN-based nature of DROID-SLAM's architecture, adopting a structured pruning approach with fine-tuning is not straightforward. The specific layer-wise sensitivity analysis-based pruning, along with an iterative fine-tuning framework, has been implemented for the DROID-SLAM architecture, as described in Section III. 

\subsection{Quantization}
Quantization in DNN optimization reduces the precision of numerical data from high-precision floating-point to lower-precision formats, usually to 8-bit integers. This method effectively addresses the objectives of minimizing storage requirements and computational overhead, thereby facilitating the deployment of quantized DNNs on 8-bit hardware platforms, including embedded systems\cite{jacob2018quantization}. 

There are two primary quantization strategies: Quantization Aware Training (QAT) and Post-Training Quantization (PTQ). QAT \cite{hanif2022crosslayer} \cite{jacob2018quantization} \cite{jain2019trained} involves retraining the network to offset accuracy losses due to quantization, often yielding improved results. PTQ \cite{banner2018posttraining} \cite{cai2020zeroq} \cite{nagel2020upordown}, on the other hand, is a simpler method that applies quantization to an existing model without further training, utilizing a representative dataset to determine quantization parameters. 

Expanding the scope of quantization to large language models (LLMs), techniques such as low-bit weight and activation quantization have proven to be crucial in dramatically enhancing LLM deployment efficiency on hardware-limited devices. These approaches, as demonstrated in recent studies \cite{zhao2023atom} \cite{dettmers2023qlora}, significantly boost serving throughput and reduce memory consumption without compromising model accuracy. Moreover, in the vision domain, PTQ has been applied to a variety of CNN architectures including Mobilenet-V1, Mobilenet-V2, Nasnet-Mobile, Inception-V3, Resnet-V1-50, Resnet-V1-152, Resnet-V2-50, and Resnet-V2-152 \cite{krishnamoorthi2018quantizing}. 

We perform PTQ to make the model 8-bit integer hardware compatible for efficient deployment, detailed in Section III.


\section{OUR METHODOLOGY}

\subsection{Structured Pruning with Fine-tuning}

Our approach to pruning DNN layers of the DROID-SLAM architecture relies on the principles of layer-wise sensitivity analysis, extending beyond simple saliency-based pruning, as developed in the paper [22]. Unlike the method applied in the paper, which involves sensitivity analysis at every pruning step, we conduct sensitivity analysis only at two mid intervals of the desired pruning percentage. This modification is made given the dense architecture of DROID-SLAM, significantly reducing the computational load and time.


\textbf{Sensitivity analysis: } For sensitivity analysis (see Fig \ref{fig:sensitivity_combined}, Phase 1), iterate over all the DNN layers:

\begin{enumerate}
    \item For a selected layer \( L_i \), compute the filter's saliency using an L1-norm metric.
    
    \item Based on the saliency measures, prune x\% of the least significant filters, creating a pruned model \( M(L_i) \). 
    
    \item The pruned model is tested on all monocular sequences of the EuRoC dataset, with the induced error recorded to assess the effects of layer pruning.
        
    \item Compute the relative sensitivity \(S(L_i)\) as the ratio of the error induced by pruning that layer \(E(L_i)\) to the total error induced by all pruned models \(E_T\).
    
    \item Get parameter fraction \( F(L_i) \) for each layer \( L_i \).
\end{enumerate}

\textbf{Pruning with iterative-finetuning: } Following sensitivity analysis, structured pruning with iterative fine-tuning is performed (see Fig \ref{fig:sensitivity_combined}, Phase 2), given a user-defined global pruning percentage \( P \). 

\begin{enumerate}

    \item During the pruning process, the percentage of filters to be removed from \( L_i \) is calculated by normalizing the product of the layer's parameter \( F(L_i) \) and its relative sensitivity \( S(L_i) \) with the global pruning rate \( P_g \).
    
    \item Prune the model and perform fine-tuning using the training dataset after each pruning step. 

    \item After the last step of fine-tuning, save the model.
\end{enumerate}

\begin{figure*}[h!]
\centering
\includegraphics[width=0.85\textwidth]{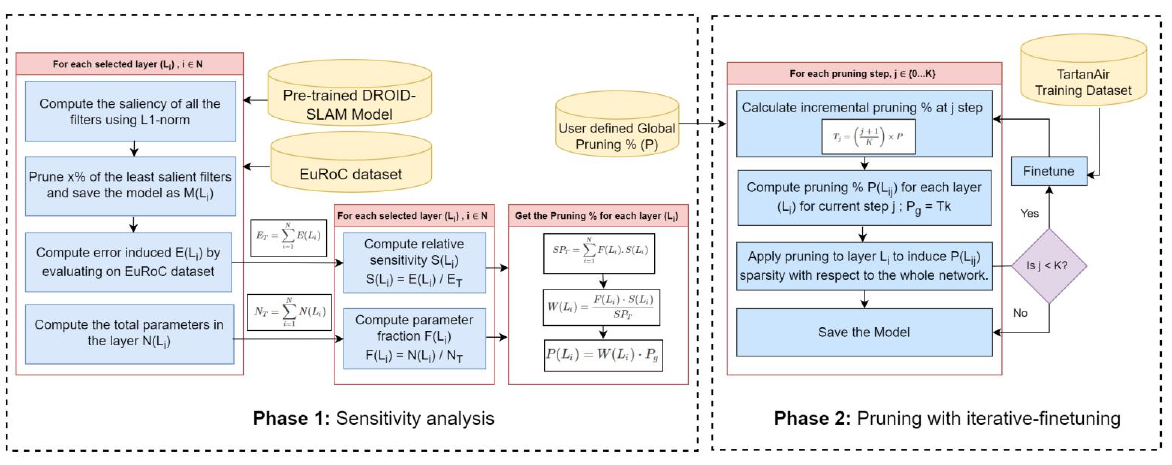}
\vspace{-10pt}
\caption{SPAQ-DL-SLAM Pruning Pipeline: Sensitivity Analysis and Pruning with Iterative-Finetuning\vspace{-10pt}}
\label{fig:pruning}
\end{figure*}

\subsection{Post-Training Quantization (PTQ)}
Following the pruning phase, the DROID-SLAM architecture undergoes post-training quantization (PTQ) to convert the model's 32-bit floating-point into 8-bit integers. 

\begin{figure}[H]
\centering
\includegraphics[width=0.25\textwidth,height=0.2\textwidth]{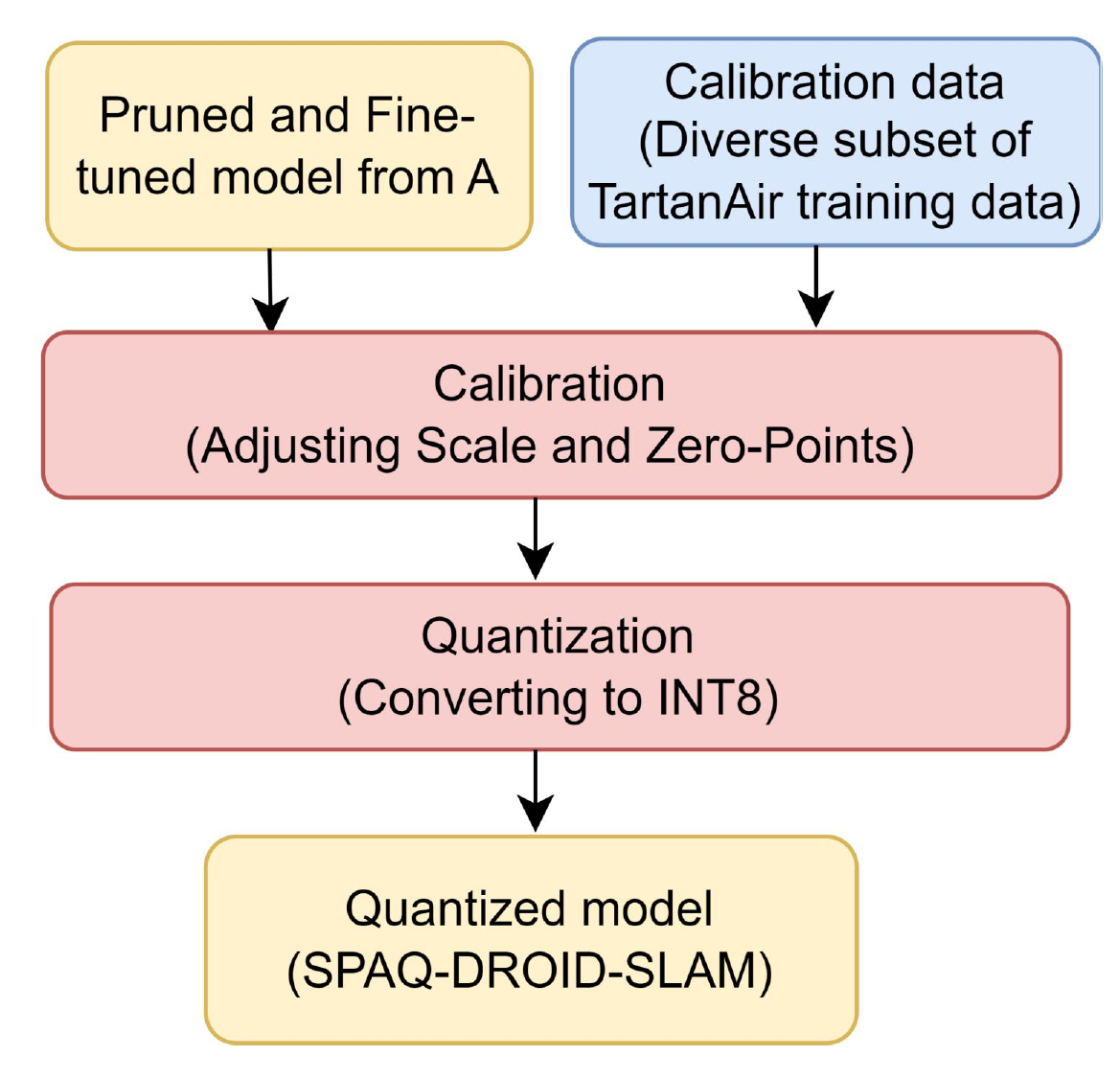}
\caption{PTQ pipeline after Cross-layer structured pruning}
\label{fig:quantizaiton_met}
\end{figure}

The PTQ process, as illustrated in Figure \ref{fig:quantizaiton_met}, begins with the initial conversion of the model to quantize both weights and activations. This is followed by a  calibration step that  fine-tunes scale and zero-point values for each layer, ensuring the quantized model preserving original accuracy as possible. This method aligns with established optimization practices highlighted in the works by \cite{jacob2018quantization} and \cite{krishnamoorthi2018quantizing}.

A subset of training TartanAir data is used as calibration data, representing the operational input distribution expected in deployment. This selection ensures that the scales and zero points are fine-tuned based on realistic scenarios. The adjustment of scale and zero points is a process that aligns the dynamic range of the quantized data with that of the original floating-point to minimize information loss \cite{hanif2022crosslayer}.

%

\section{RESULTS AND DISCUSSION}

\subsection{Evaluation Metrics and Datasets}
In evaluating SPAQ-DROID-SLAM models, we use the ATE metric, which is widely used as a standard for assessing SLAM system performance \cite{6385773}.ATE mezasures the global consistency of the estimated trajectory by computing the Root-Mean-Square Error (RMSE) between the estimated and ground truth camera poses. 


We utilize three prominent datasets for SLAM research:

    \textbf{EuRoC dataset} \cite{Burri2016TheEM}: This dataset consists of sequences captured by a Micro Aerial Vehicle (MAV), presenting challenges from slow to aggressive flight patterns. The Machine Hall datasets evaluate SLAM system adaptability to variable lighting, while the Vicon Room datasets emphasize precision in multi-view reconstruction, focusing specifically on data taken from MAVs with higher angular velocities. 

    \textbf{TUM RGB-D dataset} \cite{6385773}: This dataset offers a collection of RGB-D sequences in indoor environments, captured using a Microsoft Kinect sensor. It provides sequences with hand-held camera motions, simulating typical robot or augmented reality device movements.

    \textbf{ETH3D dataset} \cite{Schops_2019_CVPR}: This dataset presents a diverse range of indoor and outdoor scenes. It is particularly useful for evaluating SLAM systems' ability to handle complex geometries and varying scale environments.


\subsection{Sensitivity Analysis}
\begin{figure*}[h!]
\centering
\begin{minipage}{0.5\textwidth}
  \centering
  \includegraphics[width=\linewidth]{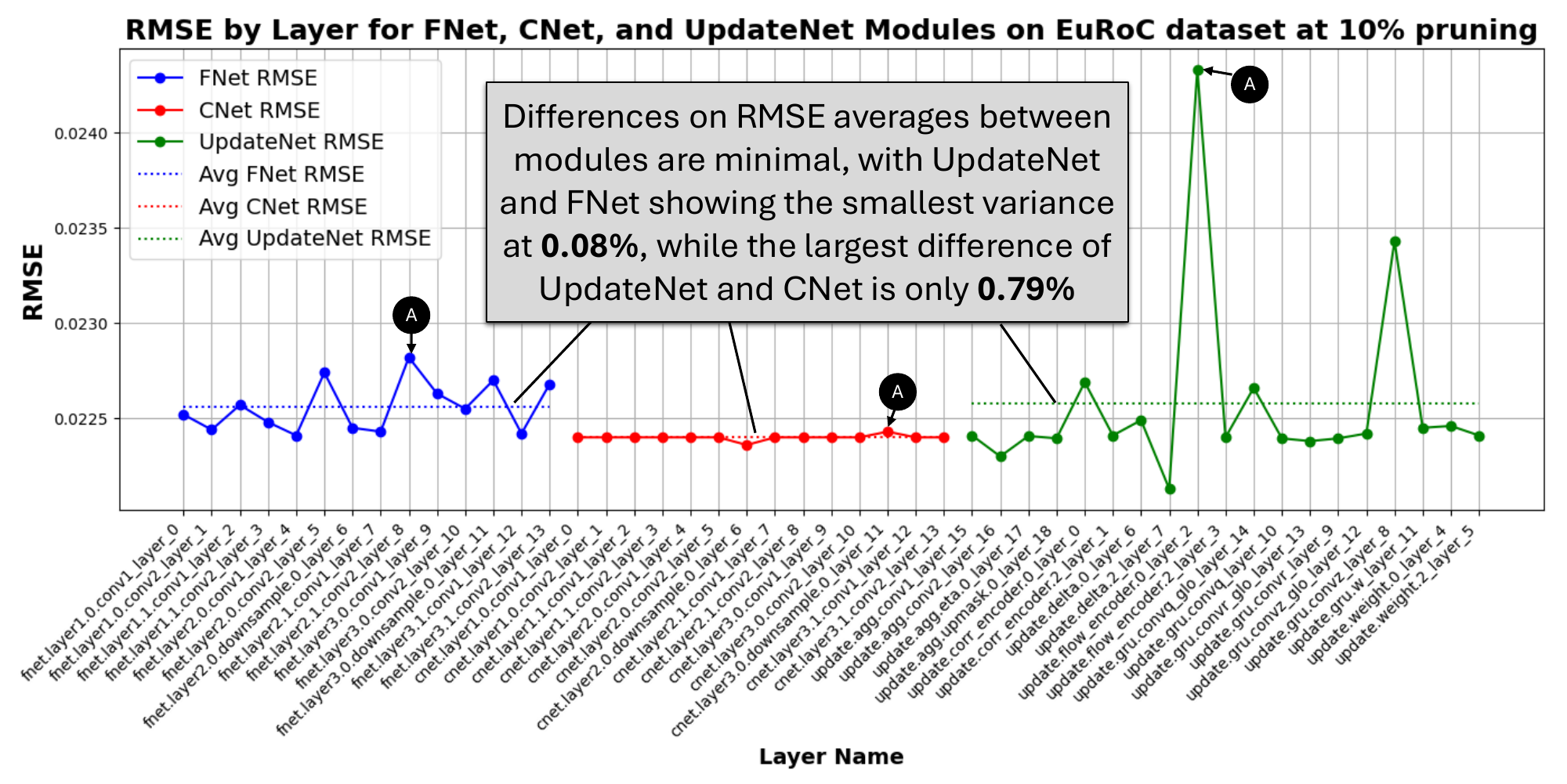}
\end{minipage}\hfill
\begin{minipage}{0.5\textwidth}
  \centering
  \includegraphics[width=\linewidth]{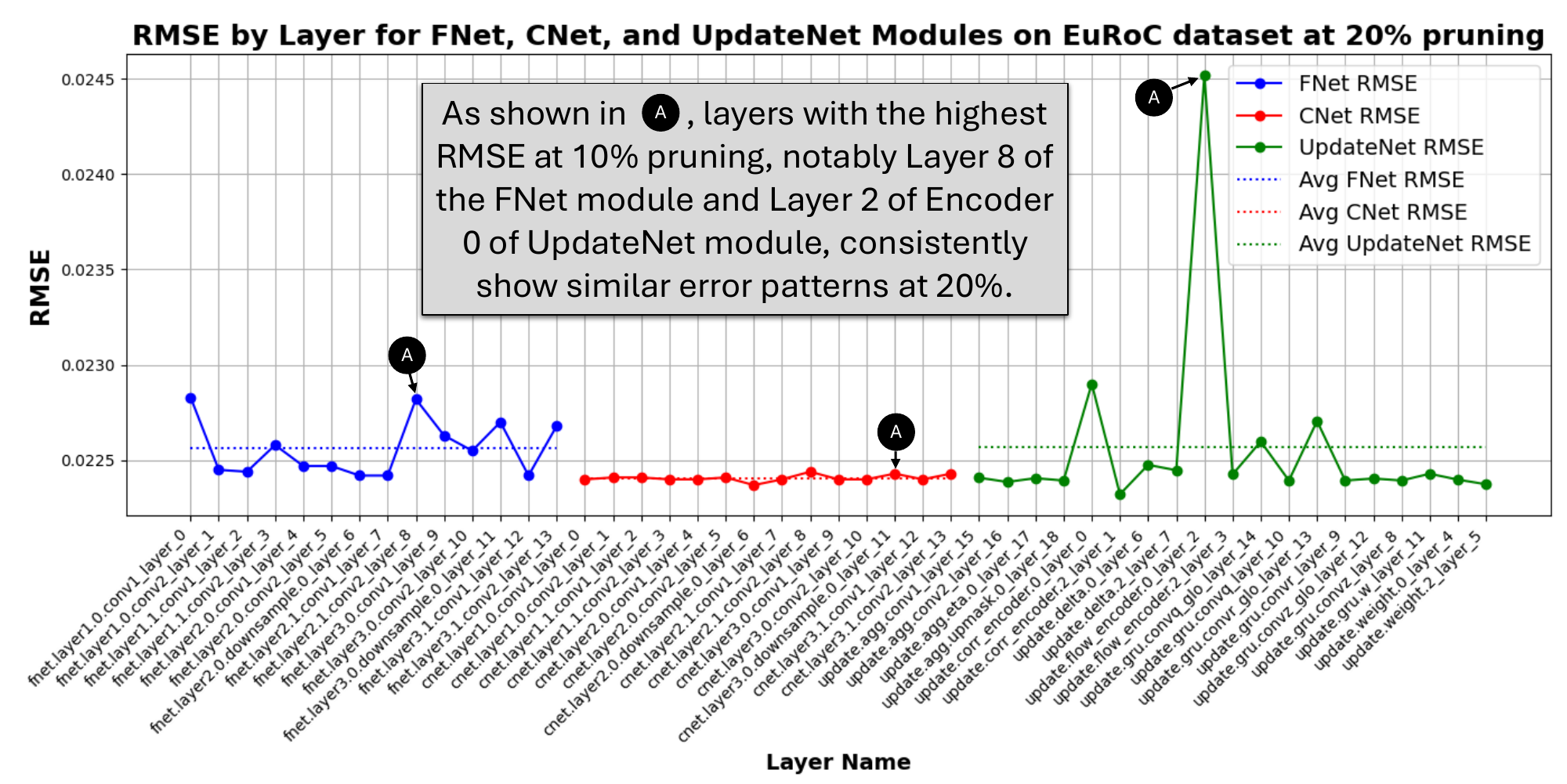}
\end{minipage}
\caption{Sensitivity analysis of FNet, CNet, and UpdateNet modules at 10\% and 20\% pruning.}
\label{fig:sensitivity_combined}
\end{figure*}

The sensitivity analysis shows that layers with the highest errors at 10\% pruning have similar pattern at 20\% pruning (see Fig \ref{fig:sensitivity_combined} \raisebox{-0.11\height}{\includegraphics[height=2.2ex]{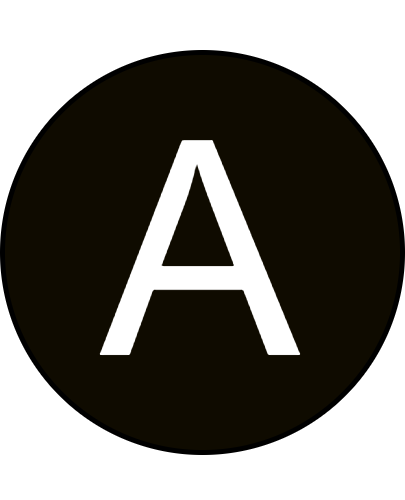}}). Notably, critical layers such as Layer 8 of the FNet module, and Layer 2 of Encoder 0 module, maintain their significance across varying pruning. This confirms that the layers identified as sensitive retain their importance after subsequent pruning iteration, irrespective of the sparsity.

Moreover, the differences in average RMSE across the FNet,CNet, and UpdateNet modules are minimal. Specifically, the variance between UpdateNet and FNet is the smallest at just 0.08\%, while the largest observed difference, between UpdateNet and CNet, is merely 0.79\%. Given this uniformity, modular-wise attention while performing pruning for DROID-SLAM's architecture turned out to be irrelevant.

\subsection{Results of the SPAQ-DROID-SLAM Models}
We present SPAQ-DROID-SLAM models with 10\% and 20\% pruning rate, fine-tuned for 10k, 50k, and 100k steps. The chosen pruning thresholds are justified by additional results (see Table \ref{tab:combined_datasets_result} and \ref{fig:test} in Appendix). 
\begin{table}[h!]
\centering
\caption{Result of Optimization in FLOPs and Model Size Reduction}
\label{Computation_Table}
\scalebox{0.7}{ 
\begin{tabular}{
l
S[table-format=4.2]
S[table-format=4.2]
S[table-format=2.2]
S[table-format=5.2]
S[table-format=5.2]
S[table-format=2.2]
}
\toprule
& \multicolumn{3}{c}{FLOPs (B)} & \multicolumn{3}{c}{Model Size (MB)} \\
\cmidrule(lr){2-4} \cmidrule(lr){5-7}
{Pruning (\%)} & {Baseline} & {Pruned} & {FLOPs ↓(\%)} & {Baseline} & {Reduced} & {Size ↓(\%)} \\
\midrule
10 & 4.64 & 4.20 & 9.44 & 15.32 & 3.63 & 76.3\\
20 & 4.64 & 3.76 & 18.90 & 15.32 & 3.25 & 79.8 \\
\bottomrule
\end{tabular}}
\end{table}

At a pruning level of 10\%, the number of floating-point operations (FLOPs) was reduced from 4.64 billion to 4.20 billion, achieving a reduction of 9.44\%. Additionally, the model size decreased from 15.32 MB to 3.63 MB, resulting in a 76.3\% reduction. When more aggressive pruning was applied at a level of 20\%, the FLOPs decreased by 18.90\%, dropping to 3.76 billion. The model size was further reduced to 3.25 MB, representing a decrease of 79.8\%.
\begin{table}[h!]
  \centering
  \caption{RMSE results on Monocular SLAM evaluation of TUM-RGBD datasets of SPAQ-DROID-SLAM models in ATE[m] metric.}
  \label{tab:tum_slam_results_quantized}
  \resizebox{\columnwidth}{!}{%
    \begin{tabular}{@{}lcccccccccc@{}}
      \toprule
      Sequence & \multicolumn{1}{c}{Original} & \multicolumn{3}{c}{10k Steps} & \multicolumn{3}{c}{50k Steps} & \multicolumn{3}{c}{100k Steps} \\
      \cmidrule(lr){2-2} \cmidrule(lr){3-5} \cmidrule(lr){6-8} \cmidrule(lr){9-11}
      Freiburg1 &  & No P & 10\% P & 20\% P & No P & 10\% P & 20\% P & No P & 10\% P & 20\% P \\
      \midrule
      360   & 0.111 & 0.093 & 0.133 & 0.069 & 0.098 & 0.079 & 0.087 & 0.089 & 0.088 & 0.078 \\
      desk  & 0.018 & 0.018 & 0.018 & 0.021 & 0.018 & 0.018 & 0.019 & 0.018 & 0.018 & 0.019 \\
      desk2 & 0.042 & 0.025 & 0.032 & 0.030 & 0.027 & 0.025 & 0.026 & 0.025 & 0.025 & 0.027 \\
      floor & 0.021 & 0.024 & 0.023 & 0.057 & 0.023 & 0.022 & 0.022 & 0.022 & 0.021 & 0.023 \\
      plant & 0.016 & 0.020 & 0.017 & 0.019 & 0.018 & 0.017 & 0.021 & 0.018 & 0.016 & 0.017 \\
      room  & 0.049 & 0.067 & 0.051 & 0.059 & 0.048 & 0.042 & 0.052 & 0.064 & 0.044 & 0.052 \\
      rpy   & 0.026 & 0.021 & 0.023 & 0.023 & 0.022 & 0.023 & 0.022 & 0.023 & 0.023 & 0.023 \\
      teddy & 0.048 & 0.034 & 0.034 & 0.038 & 0.039 & 0.034 & 0.048 & 0.034 & 0.035 & 0.063 \\
      xyz   & 0.012 & 0.010 & 0.010 & 0.010 & 0.010 & 0.010 & 0.010 & 0.010 & 0.010 & 0.010 \\
      \midrule
      Average & 0.038 & 0.035 & 0.038 & 0.036 & 0.034 & 0.030 & 0.034 & 0.034 & 0.031 & 0.034 \\
      \bottomrule
    \end{tabular}
  }
\end{table}

\begin{figure}[h!]
\centering
\includegraphics[width=0.5\textwidth]{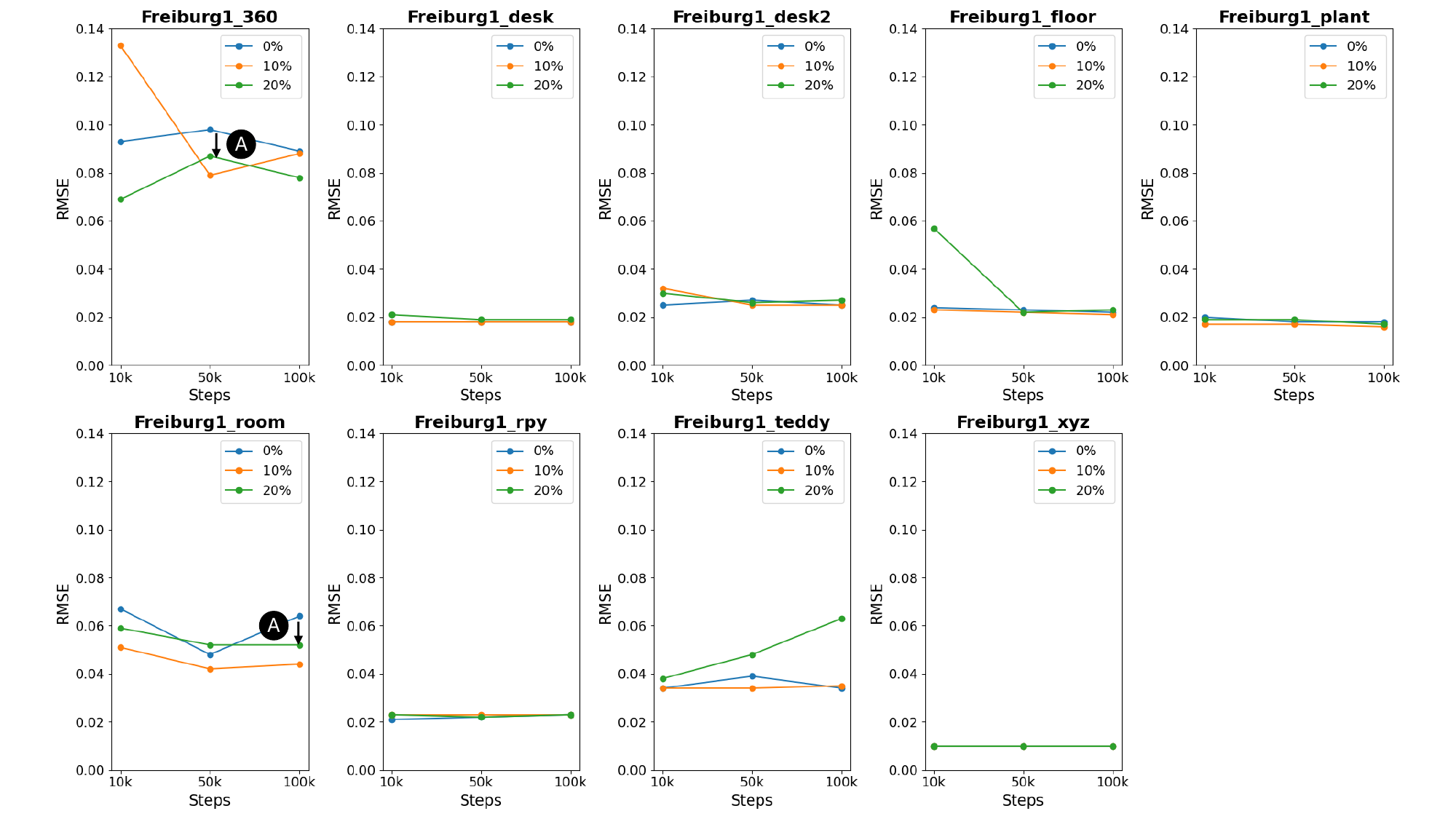}
\caption{RMSE comparison of SPAQ-DROID-SLAM models with 0\%, 10\%, and 20\% pruning at 10k, 50k, and 100k fine-tuning steps on TUM-RGBD.}
\label{fig:pruned_droidslam_rmse_comparison_tumrgbd}
\end{figure}

\textbf{TUM-RGBD evaluation: } SPAQ-DROID-SLAM models perform comparably to (see Table \ref{tab:tum_slam_results_quantized} and Fig \ref{fig:pruned_droidslam_rmse_comparison_tumrgbd}), and in some cases surpass (see Fig \ref{fig:pruned_droidslam_rmse_comparison_tumrgbd} \raisebox{-0.11\height}{\includegraphics[height=2.2ex]{Figures/label_a.png}}), the original model. At 50k fine-tuning steps, the 10\% pruned  model demonstrated a 21\% improvement in accuracy on average, while the model pruned at 20\% showed improvement of 11\% on average on ATE metric compared to the baseline DROID-SLAM. 

This suggests that the SPAQ-DL-SLAM framework, by reducing computational redundancy and focusing on the most crucial features, enhances performance. Additionally, unpruned models, noted in the 'No P' column, outperform the baseline, indicating that further training is beneficial. However, the superior results from 10\% pruned models at 50k and 100k finetuning steps demonstrate that moderate pruning within the SPAQ-DL-SLAM framework increases learning efficiency, likely improving robustness in feature extraction and error minimization for SLAM applications.


\begin{figure}[h!]
\centering
\includegraphics[width=0.5\textwidth]{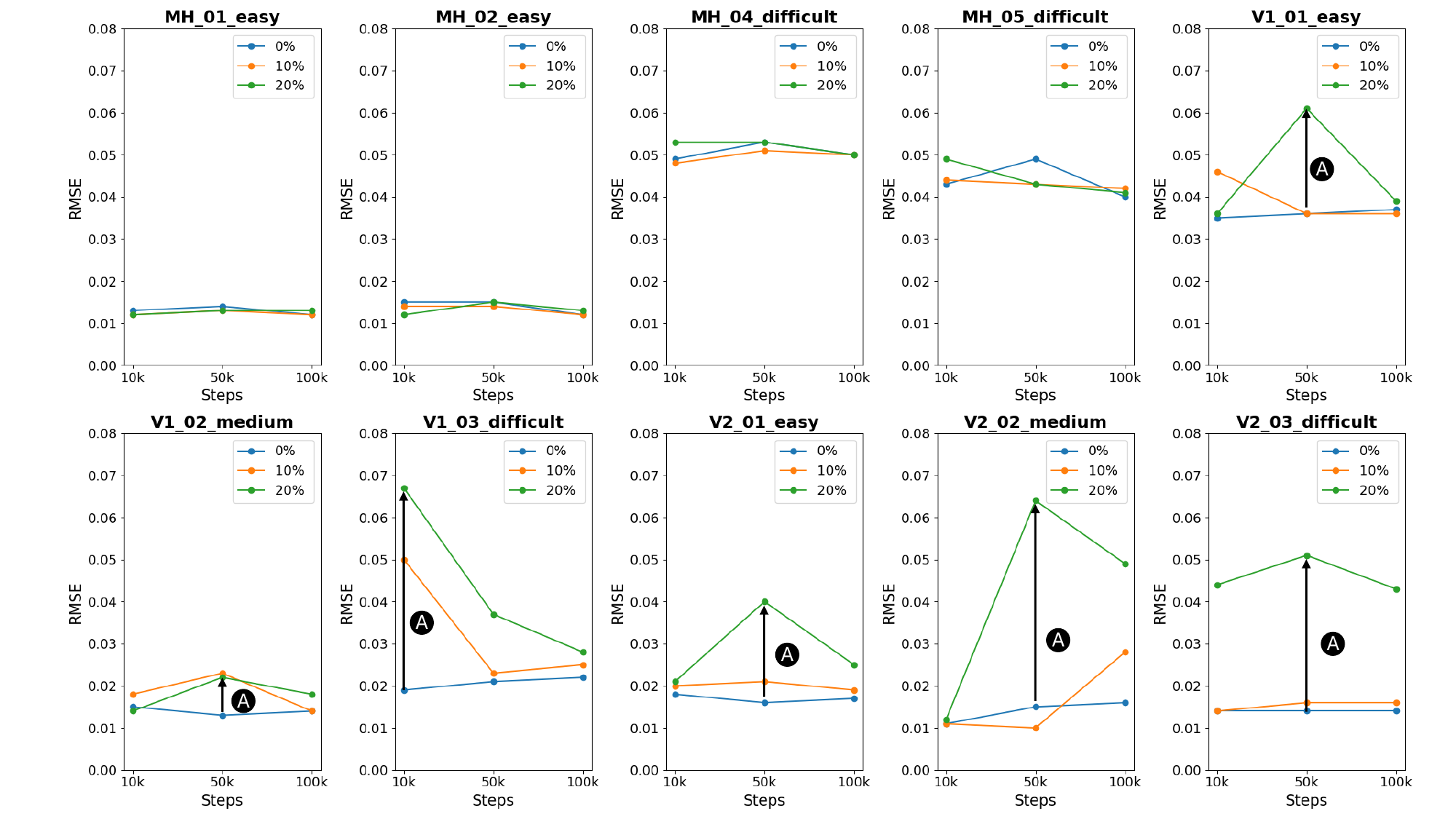}
\caption{RMSE comparison of SPAQ-DROID-SLAM models with 0\%, 10\% and 20\% pruning with 10k, 50k, and 100k fine-tuning steps on EuRoC.}
\label{fig:pruned_droidslam_rmse_comparison_euroc}
\end{figure}

\textbf{EuRoC evaluation:} SPAQ-DROID-SLAM models showed distinct effects on the Machine Hall (MH) and Vicon Room (V) datasets (see Table \ref{tab:eurorc_dataset_result} and Fig \ref{fig:pruned_droidslam_rmse_comparison_euroc}). 
\begin{table}[h!]
  \centering
  \caption{RMSE results on Monocular SLAM evaluation of EuRoC datasets of SPAQ-DROID-SLAM models in ATE[m] metric.}
  \label{tab:eurorc_dataset_result}
  \resizebox{\columnwidth}{!}{%
    \begin{tabular}{@{}lcccccccccc@{}}
      \toprule
      Sequence & Original & \multicolumn{3}{c}{10k Steps} & \multicolumn{3}{c}{50k Steps} & \multicolumn{3}{c}{100k Steps} \\
      \cmidrule(lr){3-5} \cmidrule(lr){6-8} \cmidrule(lr){9-11}
      & & No P & 10\% P & 20\% P & No P & 10\% P & 20\% P & No P & 10\% P & 20\% P \\
      \midrule
      MH01 & 0.013 & 0.013 & 0.012 & 0.012 & 0.014 & 0.013 & 0.013 & 0.012 & 0.012 & 0.013 \\
      MH02 & 0.014 & 0.015 & 0.014 & 0.012 & 0.015 & 0.014 & 0.015 & 0.012 & 0.012 & 0.013 \\
      MH04 & 0.043 & 0.049 & 0.048 & 0.053 & 0.048 & 0.051 & 0.053 & 0.050 & 0.050 & 0.050 \\
      MH05 & 0.043 & 0.043 & 0.044 & 0.049 & 0.047 & 0.043 & 0.043 & 0.040 & 0.042 & 0.041 \\
      V101 & 0.037 & 0.035 & 0.046 & 0.036 & 0.036 & 0.036 & 0.061 & 0.037 & 0.036 & 0.039 \\
      V102 & 0.012 & 0.015 & 0.018 & 0.014 & 0.013 & 0.023 & 0.022 & 0.014 & 0.014 & 0.018 \\
      V103 & 0.020 & 0.019 & 0.050 & 0.067 & 0.021 & 0.023 & 0.037 & 0.022 & 0.025 & 0.028 \\
      V201 & 0.017 & 0.018 & 0.020 & 0.021 & 0.016 & 0.021 & 0.040 & 0.017 & 0.019 & 0.025 \\
      V202 & 0.013 & 0.011 & 0.011 & 0.012 & 0.015 & 0.010 & 0.064 & 0.016 & 0.028 & 0.049 \\
      V203 & 0.014 & 0.014 & 0.014 & 0.044 & 0.014 & 0.014 & 0.051 & 0.014 & 0.014 & 0.043 \\
      \midrule
      MH Avg & 0.028 & 0.030 & 0.030 & 0.032 & 0.033 & 0.030 & 0.031 & 0.029 & 0.029 & 0.029 \\
      V Avg & 0.019 & 0.019 & 0.027 & 0.032 & 0.019 & 0.021 & 0.046 & 0.020 & 0.023 & 0.030 \\
      \midrule
      Average & 0.022 & 0.023 & 0.028 & 0.032 & 0.025 & 0.025 & 0.040 & 0.023 & 0.025 & 0.032 \\
      \bottomrule
    \end{tabular}
  }
\end{table}
For the Machine Hall datasets, SPAQ-DROID-SLAM models remained robust as the error rates increased by only 3.6\% on average at 20\% pruning and 100k fine tuning steps. Conversely, the Vicon Room datasets were sensitive to the same level of optimization, showing increase of 36\% on average (see Fig. \ref{fig:pruned_droidslam_rmse_comparison_euroc} \raisebox{-0.11\height}{\includegraphics[height=2.2ex]{Figures/label_a.png}}). The Vicon Room datasets were generated from MAVs operating at rapid angular velocities, scenarios of rapid changes in viewpoint and scale.



\begin{figure}[h!]
\centering
\includegraphics[width=0.5\textwidth]{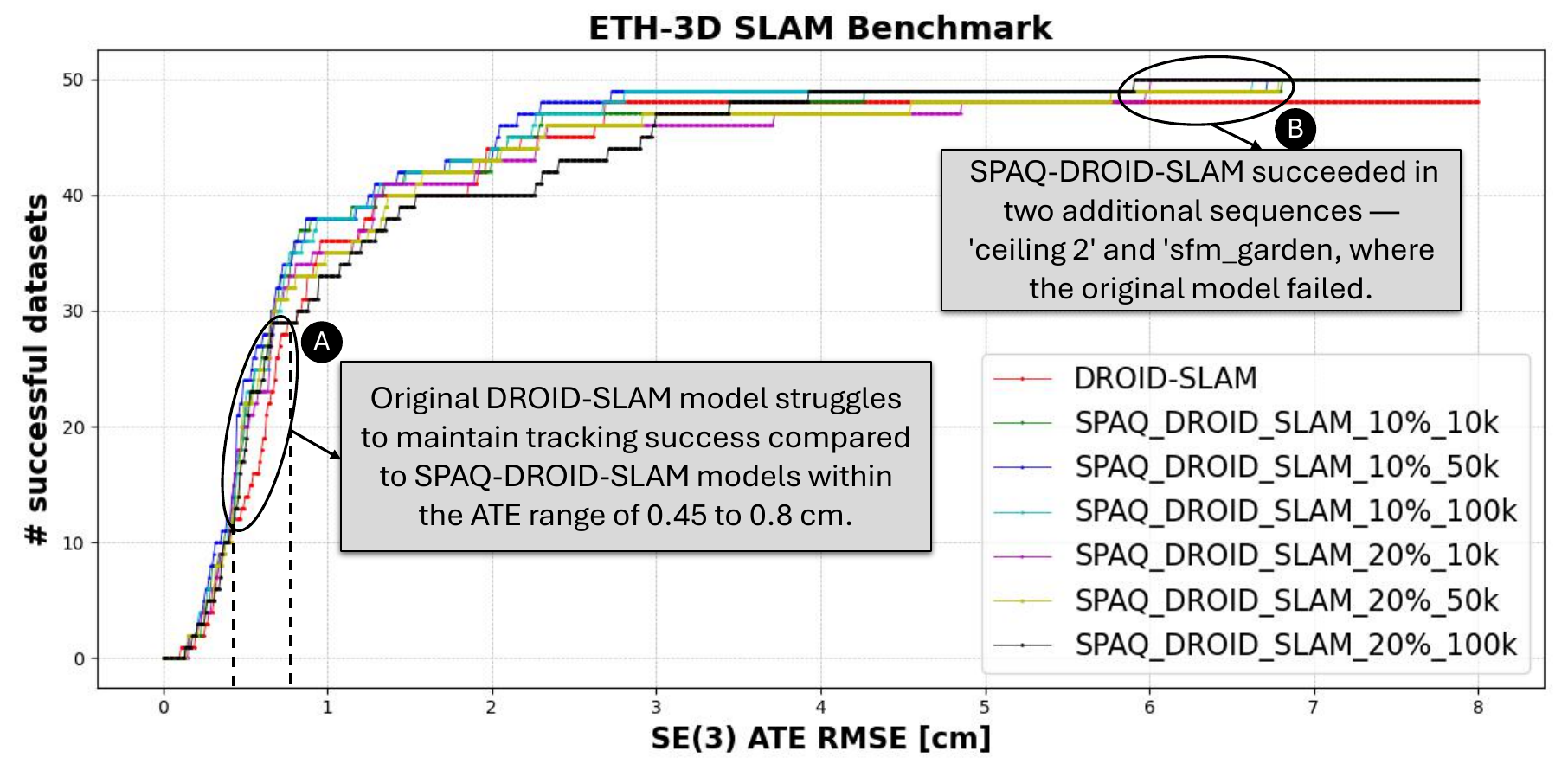} 
\caption{Comparison of the performance of SPAQ-DROID-SLAM models with the original DROID-SLAM on RGB-D ETH3D-SLAM Benchmark }
\label{fig:eth3d_slam_benchmark}
\end{figure}

\begin{table}[h!]
  \centering
  \caption{Area Under the Curve (AUC) for different methods on the ETH-3D SLAM Benchmark.}
  \label{tab:eth3d_slam_results_auc}
  \resizebox{\columnwidth}{!}{%
    \begin{tabular}{@{}lccccccc@{}}
      \toprule
      Method & DROID-SLAM & \multicolumn{3}{c}{10\% Pruned} & \multicolumn{3}{c}{20\% Pruned} \\
      \cmidrule(lr){2-2} \cmidrule(lr){3-5} \cmidrule(lr){6-8}
      Steps & Original & 10K & 50K & 100K & 10K & 50K & 100K \\
      \midrule
      AUC (train) & 340.68 & 350.54 & 355.03 & 353.01 & 344.02 & 344.64 & 342.78 \\
      \bottomrule
    \end{tabular}
  }
\end{table}

\textbf{ETH-3D SLAM evaluation:} SPAQ-DROID-SLAM models have demonstrated better performance compared to the DROID-SLAM baseline model, as measured by the Area Under the Curve (AUC) metrics (see Table \ref{tab:eth3d_slam_results_auc}). SPAQ-DROID-SLAM models clearly have better tracking success rate in the range of the ATE range of 0.45cm to 0.8cm compared to the original DROID-SLAM model (see Fig. \ref{fig:eth3d_slam_benchmark} \raisebox{-0.11\height}{\includegraphics[height=2.2ex]{Figures/label_a.png}}). In fact, all the SPAQ-DROID-SLAM models succeeded in two additional sequences —'ceiling 2' and 'sfm\_garden, where the original model failed (see Fig. \ref{fig:eth3d_slam_benchmark} \raisebox{-0.11\height}{\includegraphics[height=2.2ex]{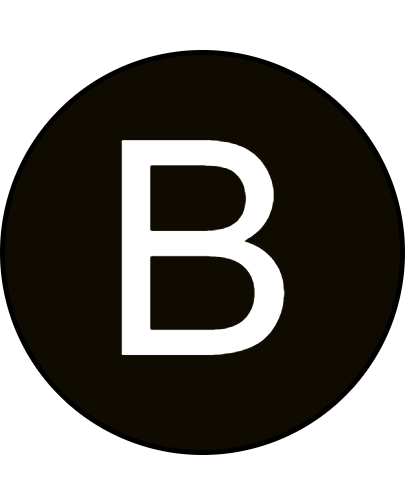}}). These results highlights enhanced generalization capabilities of SPAQ-DROID-SLAM at ETH-3D SLAM benchmark. 

Our evaluations show the distinct performance of SPAQ-DROID-SLAM  models across different datasets. While the models show enhanced capabilities on TUM RGB-D and ETH3D datasets, they encounter challenges on the Vicon Room sequences of the EuRoC dataset, characterized by high angular velocities from Micro Aerial Vehicles (MAV). These findings highlight the critical importance of considering specific operational environments and tasks in the design of DL-SLAM algorithms, which can significantly enhance both performance and resource efficiency. 


\section{CONCLUSION}
This paper introduced SPAQ-DL-SLAM, a framework implemented to optimize the DROID-SLAM algorithm for better resource and energy efficiency. Our result of SPAQ-DROID-SLAM models demonstrated improved accuracy and generalization on TUM-RGBD and ETH3D benchmarks, reducing computational demands. The performance variation observed in the distinct Vicon Room sequences highlights the need for adaptations of DL-SLAM technologies to specific environments and tasks. Our study suggests that designing DL-SLAM architectures with consideration for operational contexts facilitates efficient implementation on resource-constrained embedded platforms, enabling effective real-time on-board computation in autonomous mobile robots.
\section*{ACKNOWLEDGMENT}
This work was partially supported by the NYUAD Center for Artificial Intelligence and Robotics (CAIR), funded by Tamkeen under the NYUAD Research Institute Award CG010, and the NYUAD Center for Interacting Urban Networks (CITIES), funded by Tamkeen under the NYUAD Research Institute Award CG001.

\begin{figure*}[h!]
\centering
\includegraphics[width=0.75\linewidth]{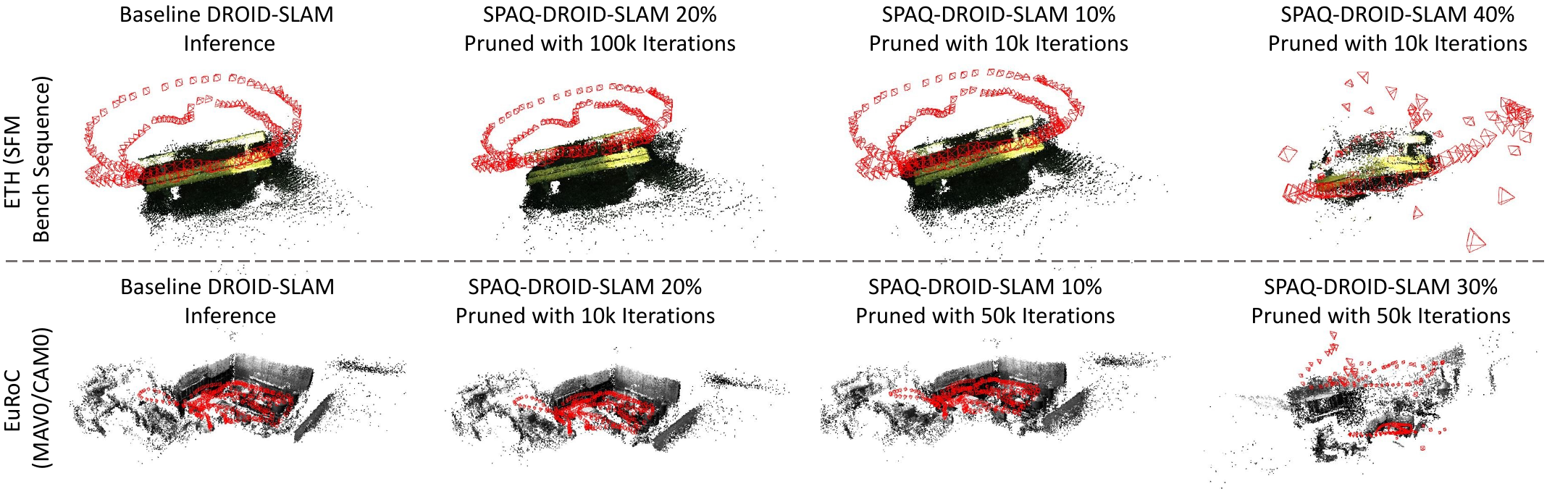}
\vspace{-10pt}
\caption{\small Comparison of 3D point cloud and poses reconstruction using DROID-SLAM and SPAQ-DROID-SLAM models.\vspace{-10pt}}
\label{fig:test}
\end{figure*}

\begin{appendix}
Additional results are presented in Table \ref{tab:combined_datasets_result}, while Figure~\ref{fig:test} illustrates the inference comparison between the original DROID-SLAM and optimized variants of SPAQ-DROID-SLAM models.

\begin{table*}[htbp]
  \centering
  \caption{Monocular SLAM evaluation of EuRoC and TUM-RGBD datasets with 25\% and 30\%, and 40\% pruning.}
  \label{tab:combined_datasets_result}
  \resizebox{0.75\textwidth}{!}{
    \begin{tabular}{@{}lcccccc|lcccccc@{}}
      \toprule
      \multicolumn{7}{c|}{\textbf{EuRoC Datasets (ATE [m])}} & \multicolumn{7}{c}{\textbf{TUM-RGBD Datasets (ATE [m])}} \\
      \midrule
      \textbf{Sequence} & \multicolumn{3}{c}{\textbf{30\% Pruning}} & \multicolumn{3}{c|}{\textbf{40\% Pruning}} & \textbf{Sequence} & \multicolumn{3}{c}{\textbf{25\% Pruning}} & \multicolumn{3}{c}{\textbf{30\% Pruning}} \\
      \cmidrule(lr){2-4} \cmidrule(lr){5-7} \cmidrule(lr){9-11} \cmidrule(lr){12-14}
      & \textbf{10k Steps} & \textbf{50k Steps} & \textbf{100k Steps} & \textbf{10k Steps} & \textbf{50k Steps} & \textbf{100k Steps} & & \textbf{10k Steps} & \textbf{50k Steps} & \textbf{100k Steps} & \textbf{10k Steps} & \textbf{50k Steps} & \textbf{100k Steps} \\
      \midrule
      MH01  & 2.301 & 5.621 & 4.451 & 4.320 & 2.244 & -  & 360     & -        & 0.211502  & 0.206238  & -        & -        & -        \\
      MH02  & 1.698 & 5.995 & 4.612 & 3.820 & 1.781 & -  & desk    & -        & 0.614238  & 0.076800  & -        & 0.082934 & 0.133804 \\
      MH04  & 1.442 & 1.771 & 1.855 & 5.177 & 5.107 & 2.567  & desk2   & -        & -         & -         & -        & -        & 0.458275 \\
      MH05  & 5.622 & 2.065 & 1.959 & -     & 4.790 & 4.996  & floor   & 0.411862 & -         & -         & -        & -        & 0.211445 \\
      V101  & 1.699 & 1.909 & 2.121 & 1.700 & 1.639 & 1.756  & plant   & -        & 0.756441  & 0.489154  & -        & -        & 0.756643 \\
      V102  & 1.943 & 2.028 & 1.733 & 2.108 & 2.028 & 2.011  & room    & -        & -         & 0.933198  & -        & 0.946653 & 0.883455 \\
      V103  & 1.442 & -     & -     & 1.936 & 1.733 & 1.485  & rpy     & 0.062179 & 0.062346  & 0.062588  & -        & 0.062420 & 0.056495 \\
      V201  & -     & -     & -     & -     & -     & 2.038  & teddy   & 0.830744 & -         & -         & -        & 0.768941 & 0.330686 \\
      V202  & -     & -     & -     & -     & -     & -      & xyz     & 0.067677 & 0.185230  & 0.127053  & -        & -        & -        \\
      V203  & -     & -     & -     & -     & -     & -      &         &          &           &           &          &          &          \\
      \bottomrule
    \end{tabular}
  }
\end{table*}

\section{Demonstration of DROID-SLAM model and SPAQ-DROID-SLAM model performances}

\end{appendix}

\FloatBarrier
\bibliography{biblio.bib}  

\end{document}